# End-to-End Waveform Utterance Enhancement for Direct Evaluation Metrics Optimization by Fully Convolutional Neural Networks

Szu-Wei Fu, Tao-Wei Wang, Yu Tsao*, Xugang Lu, and Hisashi Kawai

*Abstract*— Speech enhancement model is used to map a noisy speech to a clean speech. In the training stage, an objective function is often adopted to optimize the model parameters. However, in most studies, there is an inconsistency between the model optimization criterion and the evaluation criterion on the enhanced speech. For example, in measuring speech intelligibility, most of the evaluation metric is based on a short-time objective intelligibility (STOI) measure, while the frame based minimum mean square error (MMSE) between estimated and clean speech is widely used in optimizing the model. Due to the inconsistency, there is no guarantee that the trained model can provide optimal performance in applications. In this study, we propose an end-to-end utterance-based speech enhancement framework using fully convolutional neural networks (FCN) to reduce the gap between the model optimization and evaluation criterion. Because of the utterance-based optimization, temporal correlation information of long speech segments, or even at the entire utterance level, can be considered when perception-based objective functions are used for the direct optimization. As an example, we implement the proposed FCN enhancement framework to optimize the STOI measure. Experimental results show that the STOI of test speech is better than conventional MMSE-optimized speech due to the consistency between the training and evaluation target. Moreover, by integrating the STOI in model optimization, the intelligibility of human subjects and automatic speech recognition (ASR) system on the enhanced speech is also substantially improved compared to those generated by the MMSE criterion.

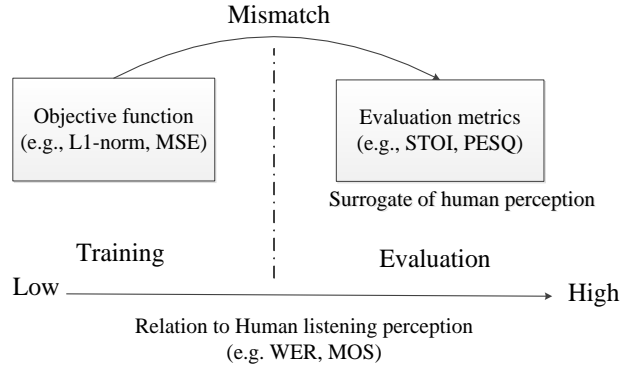

Fig. 1. Mismatch between training objective function and evaluation metrics which are usually highly correlated to human perception.

Szu-Wei Fu is with Department of Computer Science and Information Engineering, National Taiwan University, Taipei 10617, Taiwan and Research Center for Information Technology Innovation (CITI) at Academia Sinica, Taipei 11529, Taiwan (e-mail: jasonfu@citi.sinica.edu.tw).
Tao-Wei Wang is with the Research Center for Information Technology Innovation (CITI) at Academia Sinica, Taipei 11529, Taiwan (e-mail: dati1020@citi.sinica.edu.tw ).
Xugang Lu is with the National Institute of Information and Communications Technology, Tokyo 184-0015, Japan (e-mail: xugang.lu@nict.go.jp).
Hisashi Kawai is with the National Institute of Information and Communications Technology, Tokyo 184-0015, Japan (e-mail: hisashi.kawai@nict.go.jp).
Yu Tsao is with the Research Center for Information Technology Innovation (CITI) at Academia Sinica, Taipei 11529, Taiwan (e-mail: yu.tsao@citi.sinica.edu.tw ).

## I. INTRODUCTION

Recently, deep learning based spectral mapping or mask prediction frameworks for speech enhancement have been proposed and extensively investigated [1-30]. Although they were demonstrated to perform better than conventional enhancement approaches, there is still room for further improvements. For example, the objective function used for optimization in the training stage, typically the minimum mean squared error (MMSE) [31] criterion, is different from the human perception-based evaluation metrics. Formulating consistent training objectives that meet specific evaluation criteria has always been a challenging task for signal processing (generation). Since evaluation metrics are usually highly correlated to human listening perception, directly optimizing their scores may further improve the performance of enhancement model especially for the listening test. Therefore, our goal in this paper is to solve the mismatch between the objective function and the evaluation metrics as shown in Fig. 1.

For human perception, the primary goal of speech enhancement is to improve the intelligibility and quality of noisy speech [32]. To evaluate these two metrics, perceptual evaluation of speech quality (PESQ) [33] and short-time objective intelligibility (STOI) [34] have been proposed and used as objective measures by many related studies [1-5, 10-17]. However, most of them did not use these two metrics as the objective function for optimizing their models. Instead, they simply minimized the mean square error (MSE) between clean and enhanced features. Although some research [10, 11] introduced human perception into the objective function, they are still different from the final evaluation metrics. Optimizing a substitute objective function (e.g., MSE) does not guarantee good results for the true targets. We will discuss this problem and give some examples in detail in Section III.

The reasons for not directly applying the evaluation metrics as objective functions may not only be due to the complicated computation, but also because the whole (clean and processed) utterances are needed to accomplish the evaluation. Usually, conventional feed-forward deep neural networks (DNNs) [1] enhance noisy speech in a frame-wise manner due to restrictions of the model structures. In other words, during the training process, each noisy frame is individually optimized (or some may include context information). On the other hand, recurrent neural networks (RNNs) and long short-term memory (LSTM) networks, can treat an utterance as a whole and has been shown to outperform DNN-based speech enhancement models [9, 24-28]. For example, Hershey *et al.*[35] combined LSTM and global K-means on the embeddings of the whole utterance. Although LSTM may also be suitable for solving the mismatch issue between the evaluation metrics and the employed objective function, in this study, we apply the fully convolutional neural network (FCN) to perform speech enhancement in an utterance-wise manner.

An FCN model is very similar to a conventional convolutional neural network (CNN), except that the top fully connected layers are removed [36]. Therefore, it only consists of convolutional layers, and hence the local feature structures can be effectively preserved with a relatively small number of weights. Through this property, waveform-based speech enhancement by FCN was proposed, and it achieved considerable improvements when compared to DNN-based models [37]. Here, we apply another property of FCN to achieve utterance-based enhancement, even though each utterance has a different length. The reason that DNN and CNN can only process fixed-length inputs [38] is that the fully connected layer is indeed a matrix multiplication between the weight matrix and outputs of the previous layer. Because the shape of the weight matrix is fixed when the model structure (number of nodes) is decided, it is infeasible to perform multiplication on non-fixed input length. However, the filters in convolution operations can accept inputs with variable lengths.

We mainly follow the framework established in [37] to construct an utterance-based enhancement model. Based on this processing structure, we further utilize STOI as our objective function. There are three reasons why we only focus on optimizing STOI in this study. First, the computation of PESQ is much more complicated. In fact, some functions (e.g., the asymmetry factor for modeling the asymmetrical disturbance) in PESQ computation are non-continuous, so the gradient descent-based optimization cannot be directly applied [39] (this problem can be solved by substituting a continuous approximation function for the non-continuous function or by reinforcement learning, as presented in [40]). Second, improving speech intelligibility is often more challenging than enhancing quality [41, 42]. Because the MMSE criterion used in most conventional learning algorithms are not designed to directly improve intelligibility, the STOI based optimization criterion is expected to perform better. Third, some researches [43, 44] have shown that the correlation coefficient (CC) between the improvement in word error rate (WER) of ASR and the improvement in STOI is higher than other objective evaluation scores (e.g., PESQ). Their findings may suggest that a speech enhancement front-end designed by considering both MMSE and STOI may achieve better ASR performance than that by considering MMSE only. Please also note that the proposed utterance-based FCN enhancement model can handle any kind of objective functions from a local time scale (frame) to a global time scale (utterance). More specifically, our model can directly optimize the final evaluation criterion, and the STOI optimization demonstrated in this paper is just one example.

Experimental results on speech enhancement show that incorporating STOI into the objective function can improve not only the corresponding objective metric, but also the intelligibility of human subjects. In addition, it can also improve the robustness of ASR under noisy conditions, which is particularly important for real-world hands-free ASR applications, such as human-robot interactions [45].

The rest of the paper is organized as follows. Section II introduces the proposed FCN for utterance-based waveform speech enhancement. Section III details the optimization for STOI. The experimental results are evaluated in Section IV. Finally, Section V presents our discussion, and this paper is concluded in Section VI.

II. END-TO-END WAVEFORM BASED SPEECH ENHANCEMENT

In addition to frame-wise processing, the conventional DNN-based enhancement models have two potential disadvantages. First, they focus only on processing the magnitude spectrogram, such as log-power spectra (LPS) [1], and leave the phase in its original noisy form [1-6]. However, several recent studies have revealed the importance of phase to speech quality when speech is resynthesized back into time-domain waveforms [26, 46, 47]. Second, a great deal of pre-processing (e.g., framing, discrete Fourier transform (DFT)) and post-processing (e.g., overlap-add method, inverse discrete Fourier transform) are necessary for mapping between the time and frequency domains, thus increasing the computational load.

Although some recent studies have taken the phase components into consideration using complex spectrograms [12-14], these methods still need to transform the waveform into the frequency domain. To solve the two issues listed above, waveform-based speech enhancement by FCN was proposed and achieved considerable improvements when compared to the LPS-based DNN models [37]. In fact, other waveform enhancement frameworks based on generative adversarial networks (GANs) [48] and WaveNet [49, 50] were also shown to outperform conventional models. Although most of these methods have already achieved remarkable performance, they still processed the noisy waveforms in a frame-based (or chunk-based) manner. In other words, the final evaluation metrics were still not applied as the objective functions to train their models.

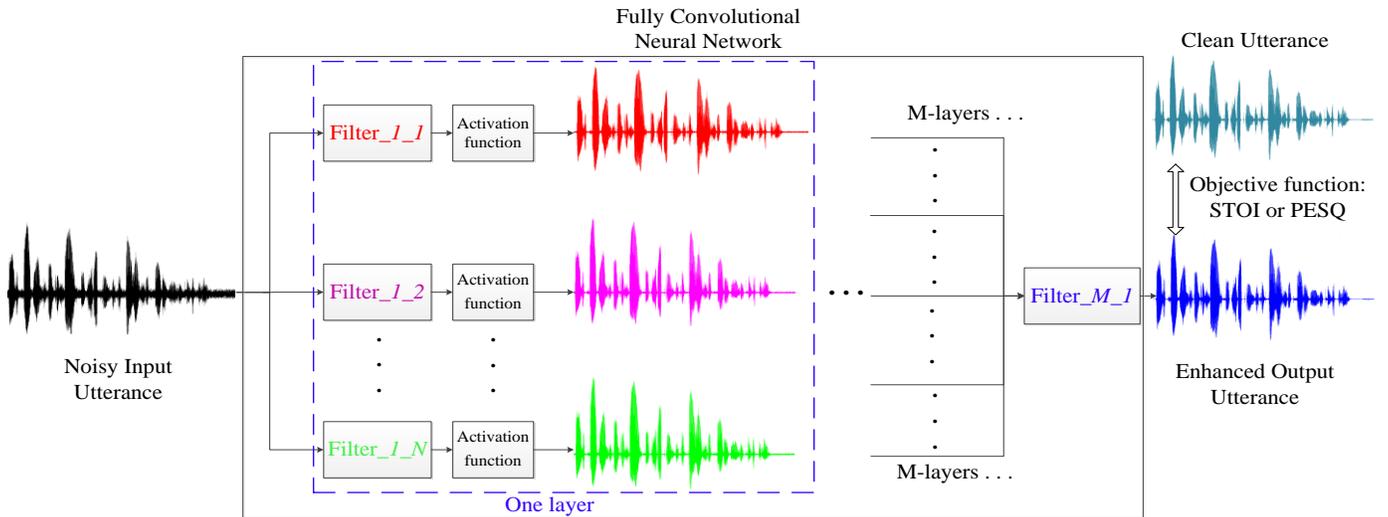

Fig. 2. Utterance-based raw waveform enhancement by FCN.

## A. FCN for Waveform Enhancement

As introduced in Introduction Section, the FCN only consists of convolutional layers; hence, the local structures of features can be effectively preserved with a relatively small number of weights. In addition, the effect of convolving a time-domain signal, $x(t)$, with a filter, $h(t)$, is equivalent to multiplying its frequency representation, $X(f)$, with the frequency response $H(f)$ of the filter [51]. Therefore, it provides some theoretical bases for FCN-based speech waveform generation.

The characteristics of a signal represented in the time domain are very different from those in the frequency domain. In the frequency domain, the value of a feature (frequency bin) represents the energy of the corresponding frequency component. However, in the time domain, a feature (sample point) alone does not carry much information; it is the relation with its neighbors that represents the concept of frequency. Fu *et al.* pointed out that this interdependency may make DNN laborious for modeling waveforms, because the relation between features is removed after fully connected layers [37]. On the other hand, because each output sample in FCN depends locally on the neighboring input regions [52], the relation between features can be well preserved. Therefore, FCN is more suitable than DNN for waveform-based speech enhancement, which has been confirmed by the experimental results in [36].

## B. Utterance-based Enhancement

In spite of the fact that the noisy waveform can be successfully denoised by FCN [37], it is still processed in a frame-wise manner (each frame contains 512 sample points). In addition to the problem of a greedy strategy [53], this also makes the convolution results inaccurate because of the zero-padding in the frame boundary. In this study, we apply another property of FCN to achieve utterance-based enhancement, even though utterances to process may have different lengths. Since all the fully connected layers are removed in FCN, the length of input features does not have to be fixed for matrix multiplication. On the other hand, the filters in the convolution operations can process inputs with different lengths. Specifically, if the filter length is $l$ and the length of input signal is $L$ (without padding), then the length of the filtered output is $L-l+1$. Because FCN only consists of convolutional layers, it can process the whole utterance without pre-processing into fixed-length frames.

Fig. 2 shows the structure of overall proposed FCN for utterance-based waveform enhancement, where Filter_*m_n* represents the *n*th filter in layer *m*. Each filter convolves with all the generated waveforms from the previous layer and produces one further filtered waveform utterance. (Therefore, filters have another dimension in the channel axis.) Since the target of (single channel) speech enhancement is to generate one clean utterance, there is only one filter, Filter_*M_1*, in the last layer. Note that this is a complete end-to-end (noisy waveform utterance in and clean waveform utterance out) framework, and there is no pre- or post-processing needed.

## III. OPTIMIZATION FOR SPEECH INTELLIGIBILITY

Several algorithms have been proposed to improve speech intelligibility based on signal processing techniques [54-56]. However, most of these algorithms focus on the applications in communication systems or multi-microphone scenarios, rather than in single channel speech enhancement, which is the main target of this paper. In addition to solving the frame boundary problem caused by zero-padding, another benefit of utterance-based optimization is the ability to design an objective function that is used for the whole utterance. In other words, each utterance is treated as a whole so that the global optimal solution (for the utterance) can be more easily obtained. Before introducing the objective function used for speech intelligibility optimization, we first show that only minimizing the MSE between clean and enhanced features may not be the most suitable target due to the characteristics of human hearing.

### A. Problems of Applying MSE as an Objective Function

One of the most intuitive objective functions used in speech enhancement is the MSE between the clean and enhanced speech. However, MSE simply compares the similarity between two signals and does not consider human perception. For example, Loizou *et al.* pointed out that MSE pays no attention to positive or negative differences between the clean and estimated spectra [41, 42]. A positive difference would signify attenuation distortions, while a negative spectral difference would signify amplification distortions. The perceptual effect of these two distortions on speech intelligibility cannot be assumed to be equivalent. In other words, MSE is not a good performance indicator of speech, and hence it is not guaranteed that better-enhanced speech can be obtained by simply minimizing MSE. The upper row of Fig. 3 shows an example of this case in the frequency domain. Although the MSE (between clean LPS and enhanced LPS) of enhanced speech in Fig. 3 (b) is lower than that in Fig. 3 (c), its performance (in terms of STOI, PESQ, and human perception) is worse than the latter. This is because the larger MSE in Fig. 3(c) results from the noisy region (highlighted in the black rectangle), which belongs to silent regions of the corresponding clean counterpart and has limited effects on the STOI/PESQ estimation. On the other hand, the spectrogram in Fig. 3 (b) is over-smoothing, and details of the speech components are missing. As pointed out in [48], the prediction results of MMSE usually bias towards an average of all the possible predictions. The two spectrograms are actually obtained from the same model, but with a different training epoch. Fig. 3 (b) is from an optimal training epoch by early stopping [57] while Fig. 3 (c) comes from an "overfitting" model due to overtraining. Note that here we use double quotes to emphasize that this overfitting is relative to the MSE criterion, and not to our true targets of speech enhancement.

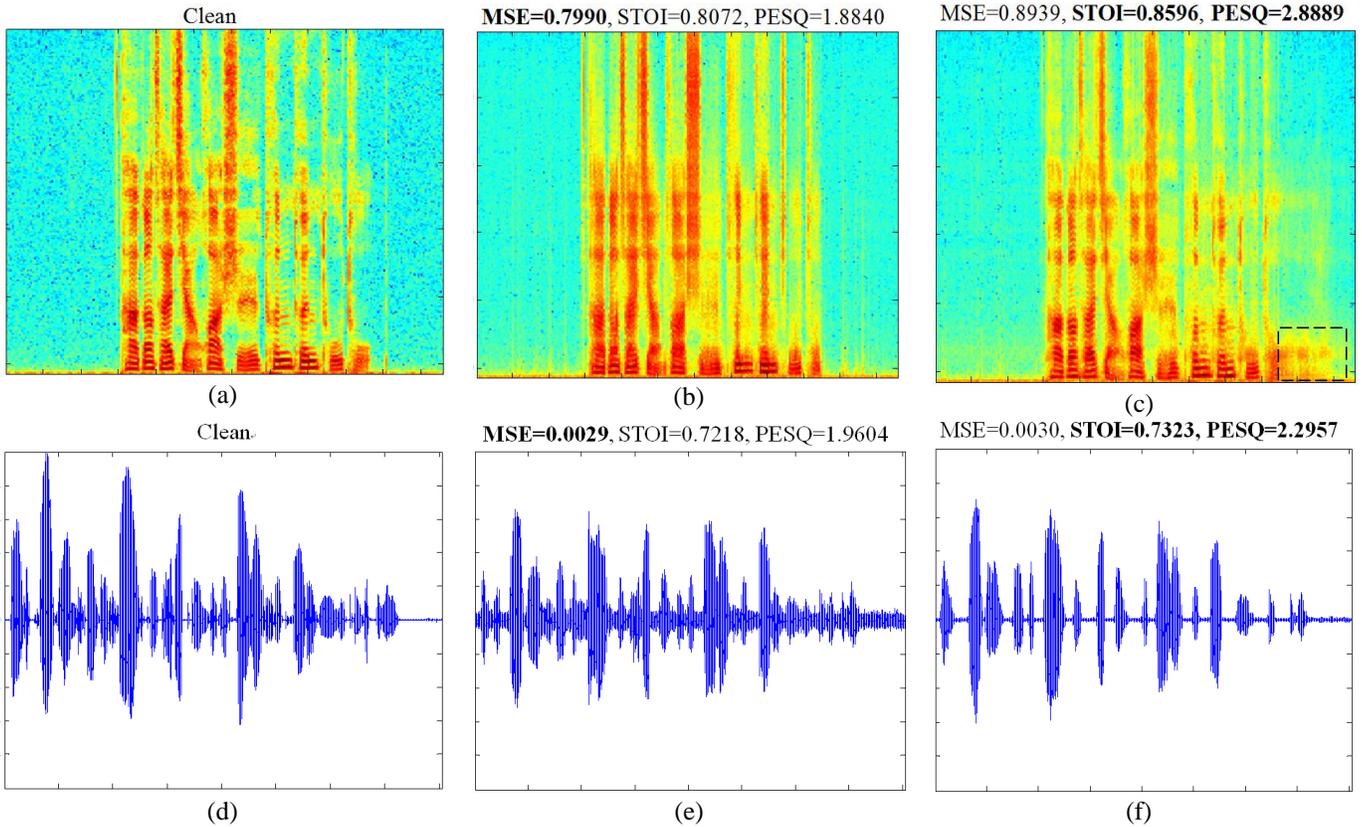

Fig. 3. An enhanced speech with lower MSE does not guarantee a better performance in evaluation. The upper row shows the case in the frequency domain, where the MSE is measured between a clean LPS and an enhanced LPS. The lower row shows the case in the time domain, where the MSE is measured between a clean waveform and an enhanced waveform.

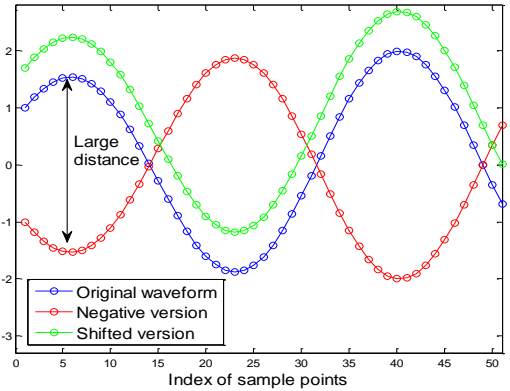

Fig. 4. The original waveform, its negative version, and its amplitude shifted version sound completely the same to humans, but the MSE between the sample points of these sounds is very large.

The above discussion implies that minimizing the MSE may make the estimated speech looks like the clean one; however, sometimes a larger MSE in the optimization process can produce speech sounds more similar to the clean version[1].

Although the waveform-based FCN enhancement model in [37] is optimized with an MSE objective function, it is also not the best target for the time domain waveform, because the relation between the MSE value and human perception is still

---

[1] We observe that this is not a single special case. A model that yields lower average MSE scores on the whole data set may not guarantee to give higher STOI and PESQ scores. Please note that, the experimental results reported in Section IV followed the common machine learning strategy that the optimized model is the one which can make the employed objective function minimized.

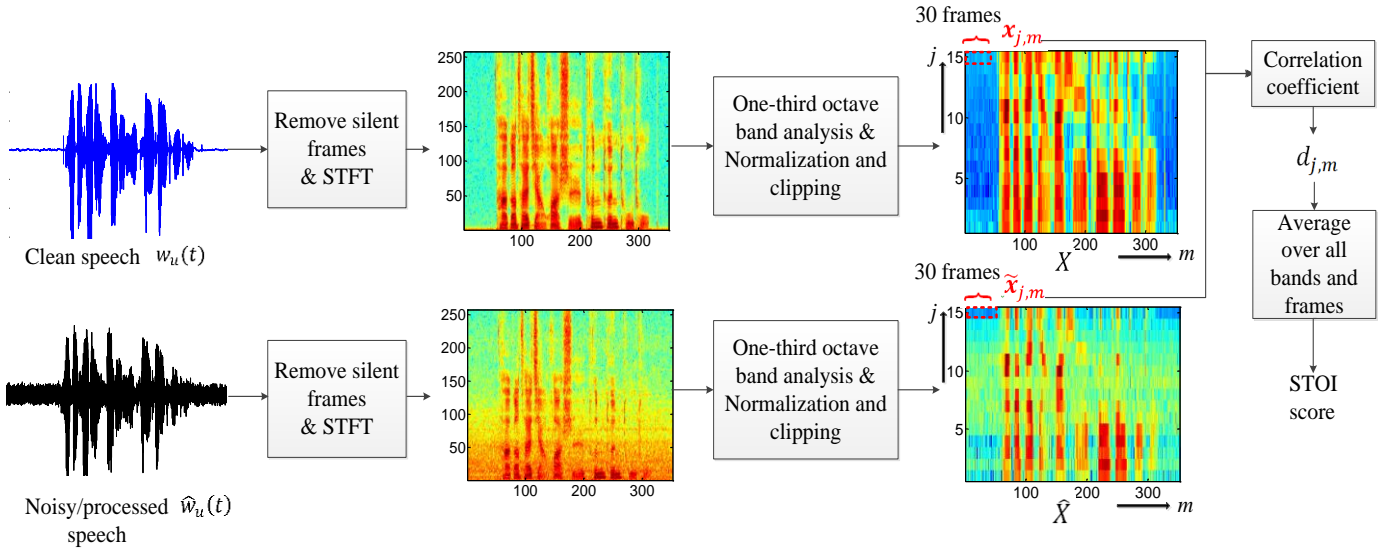
Fig. 5. Calculation of STOI is based on the correlation coefficient between the temporal envelopes of the clean and noisy/processed speech for short segments (e.g., 30 frames).

not a monotonic function. For example, as shown in Fig. 4, it is difficult for people to distinguish between a waveform, its negative version, and its amplitude shifted version by listening, although the MSE between them is very large. This also verifies the argument made in Section II-A that sample point itself does not carry much information; it is the relation with its neighbors that represent the concept of frequency. The lower row of Fig. 3 also shows a real example in the time domain in which an enhanced speech with a lower MSE (between the clean and enhanced waveforms) does not guarantee better performance. In summary, we argue that it is not guaranteed a good performance for human listening perception can be obtained by only minimizing MSE.

### B. Introduction of STOI

To overcome the aforementioned problem of MSE, here we introduce an objective function, which considers human hearing perception. The STOI score is a prevalent measure used to predict the intelligibility of noisy or processed speech. The STOI score ranges from 0 to 1, and is expected to be monotonically related to the average intelligibility of various listening tests. Hence, a higher STOI value indicates better speech intelligibility.

STOI is a function of the clean and degraded speech, and the overall computational process is illustrated as in Fig. 5. The calculation of STOI includes 5 major steps, briefly described as follows:

*1) Remove silent frames:* Since silent regions do not contribute to speech intelligibility, they are removed before evaluation.

*2) Short-time Fourier transform (STFT):* Both signals are TF-decomposed in order to obtain a representation similar to the speech representation properties in the auditory system. This is obtained by segmenting both signals into 50% overlapping Hann-windowed frames, with a length of 256 samples, where each frame is zero-padded up to 512 samples.

*3) One-third octave band analysis:* This is performed by simply grouping DFT-bins. In total, 15 one-third octave bands are used, where the lowest center frequency is set to 150 Hz and the highest one-third octave band has a center-frequency of ~4.3 kHz. The following vector notation is used to denote the short-time temporal envelope of the clean speech:

$$x_{j,m} = [X_j(m-N+1), X_j(m-N+2), \ldots X_j(m)]^T \quad (1)$$

where $X \in R^{15*M}$ is the obtained one-third octave band, $M$ is the total number of frames in the utterance, $m$ is the index of the frame, $j \in \{1,2,\ldots 15\}$ is the index of the one-third octave band, and $N = 30$, which equals an analysis length of 384 ms. Similarly, $\hat{x}_{j,m}$ denotes the short-time temporal envelope of the degraded speech.

*4) Normalization and clipping:* The goal of the normalization procedure is to compensate for global level differences, which should not have a strong effect on speech intelligibility. The clipping procedure ensures that the sensitivity of the STOI evaluation towards one severely degraded TF-unit is upper bounded. The normalized and clipped temporal envelope of the degraded speech is denoted as $\tilde{x}_{j,m}$.

*5) Intelligibility measure:* The intermediate intelligibility measure is defined as the correlation coefficient between the two temporal envelopes:

$$d_{j,m} = \frac{(x_{j,m} - \mu_{x_{j,m}})^T (\tilde{x}_{j,m} - \mu_{\tilde{x}_{j,m}})}{\|x_{j,m} - \mu_{x_{j,m}}\|_2 \|\tilde{x}_{j,m} - \mu_{\tilde{x}_{j,m}}\|_2} \quad (2)$$

where $\|.\|_2$ represents the L2-norm, and $\mu_{(\cdot)}$ is the sample mean of the corresponding vector. Finally, STOI is calculated as the average of the intermediate intelligibility measure over all bands and frames:

$$\text{STOI} = \frac{1}{15M} \sum_{j,m} d_{j,m} \quad (3)$$

The calculation of STOI is based on the correlation coefficient between the temporal envelopes of the clean and the noisy/processed speech for short segments (e.g., 30 frames). Therefore, this measure cannot be optimized by a traditional frame-wise enhancement scheme. For a more detailed setting of each step, please refer to [34].

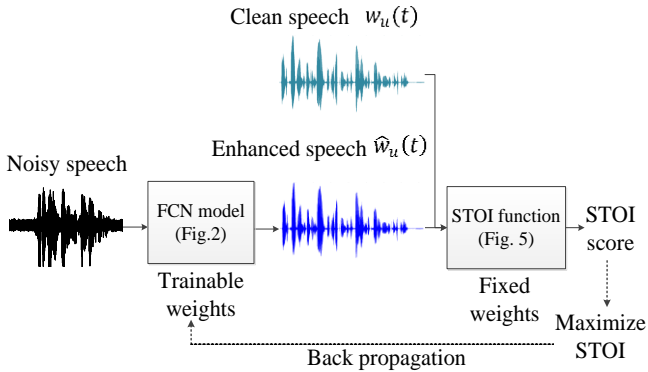

Fig. 6. The STOI computation function (Fig. 5) is cascaded after the proposed FCN model (Fig. 2) as the objective function.

TABLE I
PERFORMANCE COMPARISON OF THE TIMIT DATA SET WITH RESPECT TO STOI AND PESQ

| SNR (dB) | Frame-based [37] FCN (obj=MMSE) | | Utterance-based FCN (obj=MMSE) | | Utterance-based FCN (obj=STOI) | |
|---|---|---|---|---|---|---|
| | STOI | PESQ | STOI | PESQ | STOI | PESQ |
| 12 | 0.874 | 2.718 | 0.909 | **2.909** | **0.931** | 2.587 |
| 6 | 0.833 | 2.346 | 0.864 | **2.481** | **0.888** | 2.205 |
| 0 | 0.758 | 1.995 | 0.780 | **2.078** | **0.814** | 1.877 |
| -6 | 0.639 | 1.719 | 0.647 | **1.754** | **0.699** | 1.608 |
| -12 | 0.506 | 1.535 | 0.496 | **1.536** | **0.562** | 1.434 |
| Avg. | 0.722 | 2.063 | 0.739 | **2.152** | **0.779** | 1.942 |

## C. Maximizing STOI for Speech Intelligibility

Although the calculation of STOI is somewhat complicated, most of the computation is differentiable, and thus it can be employed as the objective function for our utterance optimization as shown in Fig. 6. Therefore, the objective function that should be minimized during the training of FCN can be represented by the following equation.

$$O = -\frac{1}{U}\sum_u stoi(w_u(t), \hat{w}_u(t)) \quad (4)$$

where $w_u(t)$ and $\hat{w}_u(t)$ are the clean and estimated utterance with index $u$, respectively, and $U$ is the total number of training utterance. $stoi(.)$ is the function that includes the five steps stated in previous section, which calculates the STOI value of the noisy/processed utterance given the clean one. Hence, the weights in FCN can be updated by gradient descent as follows:

$$f_{i,j,k}^{(n+1)} = f_{i,j,k}^{(n)} + \frac{\lambda}{B}\sum_{u=1}^{B}\frac{\partial stoi(w_u(t), \hat{w}_u(t))}{\partial \hat{w}_u(t)}\frac{\partial \hat{w}_u(t)}{\partial f_{i,j,k}^{(n)}} \quad (5)$$

Where $f_{i,j,k}^{(n+1)}$ is the $i$-th layer, $j$-th filter, $k$-th filter coefficient in FCN. $n$ is the index of the iteration number, $B$ is the batch size and $\lambda$ is the learning rate. Note that the first term in summation depends on STOI function only. We use Keras [58] and Theano [59] to perform automatic differentiation, without the need of explicitly computing the gradients of the cost function.

## IV. EXPERIMENT

In the experiment, we prepare three data sets to evaluate the performance of different enhancement models and objective functions. The first is the TIMIT corpus [60], so that the results presented here can also be compared to the frame-based FCN as reported in [37]. The second data set is the Mandarin version of the Hearing in Noise Test (MHINT) corpus [61], which is suitable for conducting listening test. The last corpus is the 2nd CHiME speech separation and recognition challenge (CHiME2) medium vocabulary track database [62], which is a more difficult challenge because it contains both additive and convolutive noise. We present the FCN model structure used in these sets of experiments in Fig. 7. Note that the frame-based FCN has the same model structure as the utterance-based FCN, except that the input is a fixed-length waveform segment (512 sample points). The comparison of frame-based FCN and LPS-based DNN are reported in our previous work [37].

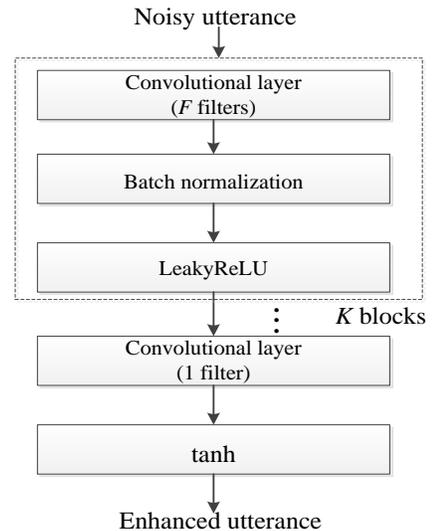

Fig. 7. The FCN structure used in this paper. In the TIMIT data set, we use $K$=5 and $F$=15 as used in [37]. In the MHINT and CHiME2 data sets, we use $K$=7 and $F$=30.

## A. Experiment on the TIMIT data set

In this set of experiments, the utterances from the TIMIT corpus were used to prepare the training and test sets. For the training set, 600 utterances were randomly selected and corrupted with five noise types (Babble, Car, Jackhammer, Pink, and Street) at five SNR levels (-10 dB, -5 dB, 0 dB, 5 dB, and 10 dB). For the test set, we randomly selected another 100 utterances (different from those used in the training set). To make the experimental conditions more realistic, both the noise types and SNR levels of the training and test sets were mismatched. Thus, we adopted three other noise signals: white Gaussian noise (WGN), which is a stationary noise, and an engine noise and a baby cry, which are non-stationary noises, using another five SNR levels (-12 dB, -6 dB, 0 dB, 6 dB, and 12 dB) to form the test set. All the results reported were averaged across the three noise types. For more detailed experiment settings and model structure, refer to [37].

To evaluate the performance of speech intelligibility, the STOI scores were used as a measure. We also present PESQ for speech quality evaluation to make a complete comparison with the results shown in [37]. (Although this metric is not optimized in this paper, we also report the results for completeness). Table

TABLE II
PERFORMANCE COMPARISON OF THE MHINT DATA SET WITH RESPECT TO STOI AND PESQ

| | Noisy | | Frame-based | | | | Utterance-based | | | | | |
| --- | --- | --- | --- | --- | --- | --- | --- | --- | --- | --- | --- | --- |
| | | | LPS | | | | Raw waveform | | | | | |
| | | | DNN (obj=MMSE) | | BLSTM (obj=MMSE) | | FCN (obj=MMSE) | | FCN (obj=STOI) | | FCN (obj=MMSE+STOI) | |
| SNR (dB) | STOI | PESQ | STOI | PESQ | STOI | PESQ | STOI | PESQ | STOI | PESQ | STOI | PESQ |
| 9 | 0.9006 | 1.744 | 0.8891 | 2.375 | 0.9052 | **2.683** | 0.9233 | 2.548 | **0.9436** | 2.306 | 0.9426 | 2.499 |
| 6 | 0.8622 | 1.554 | 0.8673 | 2.188 | 0.8875 | **2.521** | 0.9008 | 2.368 | **0.9245** | 2.115 | 0.9228 | 2.326 |
| 3 | 0.8136 | 1.383 | 0.8362 | 1.960 | 0.8600 | **2.318** | 0.8701 | 2.180 | **0.8975** | 1.902 | 0.8944 | 2.135 |
| 0 | 0.7574 | 1.238 | 0.7947 | 1.718 | 0.8236 | **2.077** | 0.8297 | 1.972 | **0.8604** | 1.656 | 0.8557 | 1.925 |
| -3 | 0.6958 | 1.102 | 0.7434 | 1.456 | 0.7718 | **1.796** | 0.7782 | 1.724 | **0.8131** | 1.388 | 0.8042 | 1.670 |
| -6 | 0.6328 | 0.945 | 0.6817 | 1.187 | 0.7128 | **1.494** | 0.7114 | 1.448 | **0.7524** | 1.131 | 0.7379 | 1.398 |
| Avg. | 0.7772 | 1.336 | 0.8020 | 1.814 | 0.8268 | **2.148** | 0.8356 | 2.040 | **0.8652** | 1.750 | 0.8596 | 1.992 |
| # of parameters | None | | 1,264,757 | | 4,433,537 | | 300,931 | | | | | |

I presents the results of the average STOI and PESQ scores on the test set for the frame-based FCN [37] and the proposed utterance-based FCN with different objective functions, where obj represents the objective function used for training. Please note that all three models have the same structure, and the only difference between them is the objective function or input unit (frame or utterance). From this table, we can see that the utterance-based FCN (with MSE objective function) can outperform frame-based FCN in terms of both PESQ and STOI. This improvement mainly comes from solving the frame boundary problem in the frame-based optimization. When employing the STOI as the objective function, it can considerably increase the STOI value (with an improvement of 0.04 on average), especially in low-SNR conditions. Although the average PESQ decreases, the STOI is enhanced, which is the main goal of this study.

*B. Experiment on the MHINT data set*
*1) Experiment Setup*

In this set of experiments, the MHINT corpus was used to prepare the training and test sets. This corpus includes 240 utterances, and we collected another 240 utterances from the same speaker to form the complete task in this study. Each sentence in the MHINT corpus consists of 10 Chinese characters and are designed to have similar phonemic characteristics among lists [61]. Therefore, this corpus is very suitable for conducting listening test. Among these 480 utterances, 280 utterances were excerpted and corrupted with 100 noise types [63], at five SNR levels (-10 dB, -5 dB, 0 dB, 5 dB, and 10 dB) as training set. Another 140 utterances and the remaining utterances were mixed to form the test set and validation set, respectively. In this experiment, we still consider a realistic condition, where both noise types and SNR levels of the training and test sets were mismatched. Thus, we intentionally adopted three other noise signals: engine noise, white noise, and street noise, with another six SNR levels: -6 dB, -3 dB, 0 dB, 3 dB, 6 dB, and 9 dB to form the test set. All the results reported were averaged across the three noise types.

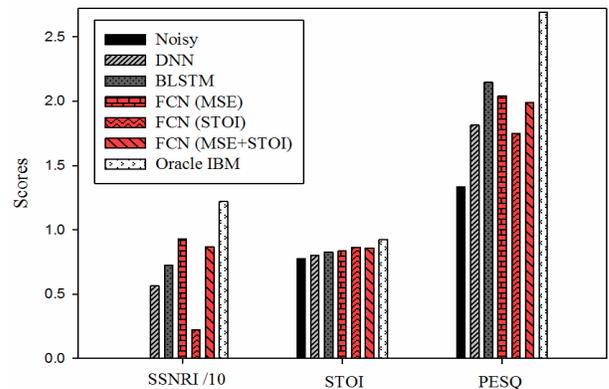
Fig. 8. Average objective evaluation scores for different models (including the oracle IBM) on the MHINT data set.

As shown in Fig. 7, the FCN model has 8 convolutional layers with zero padding to preserve the same size as the input. Except for only 1 filter used in the last layer, each of the previous layers consists of 30 filters with a filter size of 55. There are no pooling layers in the network as used in WaveNet [52]. We also train a (257 dimension) LPS-based DNN model and bidirectional long short-term memory (BLSTM) as baselines. The DNN has 5 hidden layers with 500 nodes for each layer. The BLSTM has 2 bidirectional LSTM layers, each with 384 nodes as in [26] followed by a fully connected output layer. Both the model structure and number of training epoch are decided based on monitoring the error of the validation set. Specifically, we gradually increase the number of filters, filter size, and the number of layers until the decrease of validation loss starts to saturate or the computational cost becomes intractable.

All the models employ leaky rectified linear units (LeakyReLU) [64] as the activation functions for the hidden layers. There is no activation function (linear) in the output layer of DNN and BLSTM. The FCN applies hyperbolic tangent (tanh) for output layer to restrict the range of output waveform sample points between -1 and +1. Both DNN and FCN are trained using Adam [65] optimizer with batch normalization [66]. BLSTM is trained with RMSprop [67], which is usually a suitable optimizer for RNNs.

During the STOI calculation, the first step is to exclude the silent frames (with respect to the clean reference speech). In other words, it does not consider the non-speech regions into the STOI score calculation. In addition, unlike minimizing MSE that has a unique optimal solution (i.e., for a fixed target vector $c$, the unique solution that can make MSE minimizing (equals to zero) is $c$ itself), maximizing the correlation coefficient used in (2) for intermediate intelligibility has multiple optimal solutions (i.e., for a fixed target vector $c$, the solutions that can make CC maximizing (equals to one) are $S_1 * c + S_2$. Where $S_1 > 0$ and $S_2$ is an arbitrary constant). Therefore, if we do not limit the solution space, the obtained solution may not be the one we want. Specifically, $S_1$ and $S_2$ may make the STOI-optimized speech sounds noisy as shown in the next section about Spectrogram Comparison. To process the regions not considered in STOI and constrain the solution space (for noise suppression), we also try to incorporate both the MSE and STOI into the objective function, which can be represented by the following equation.

$$O = \frac{1}{U}\sum_u (\frac{\alpha}{L_u} \|w_u(t) - \widehat{w}_u(t)\|_2^2 - stoi(w_u(t), \widehat{w}_u(t))), \quad (6)$$

where $L_u$ is the length of $w_u(t)$ (note that each utterance has a different length), and $\alpha$ is the weighting factor of the two targets. Here, $\alpha$ is simply set to 100 to balance the scale of the two targets. Since the first term can be seen as related to maximizing the SNR of enhanced speech, and the second term is to maximize the STOI, the two targets in (6) can also be considered as a multi-metrics learning [14] for speech enhancement.

*2) Experiment Results of Objective Evaluation Scores*

The STOI and PESQ scores of the enhanced speech under different SNR conditions are presented in Table II. Furthermore, we also report the average segmental SNR improvement (SSNRI) [68], STOI and PESQ by different enhancement models and oracle "ideal binary mask"(IBM) [69] (simply as a reference) in Fig. 8. Please note that the SSNRI in this figure is divided by 10 to make different metrics have similar range. From these results, we can observe that BLSTM can considerably outperform the DNN baseline. For utterance-based enhancement models, our proposed FCN (with MSE objective function) has higher SSNRI and STOI scores with lower PESQ when compared to BLSTM. Moreover, the number of parameters in FCN is roughly only 7% and 23% to that in BLSTM and DNN, respectively. When changing the objective function of FCN from MSE to STOI, the STOI value of the enhanced speech can be considerably improved with a decreased PESQ score. This may be due to the FCN process the STOI-undefined region (silent and high frequency regions) in an unsuitable way (we can more easily observe this phenomenon by spectrograms of the processed speech in the next section). Optimizing both MSE and STOI simultaneously seems to strike a good balance between speech intelligibility and quality, with PESQ and SSNRI considerably improved and STOI marginally degraded compared to STOI-optimized speech.

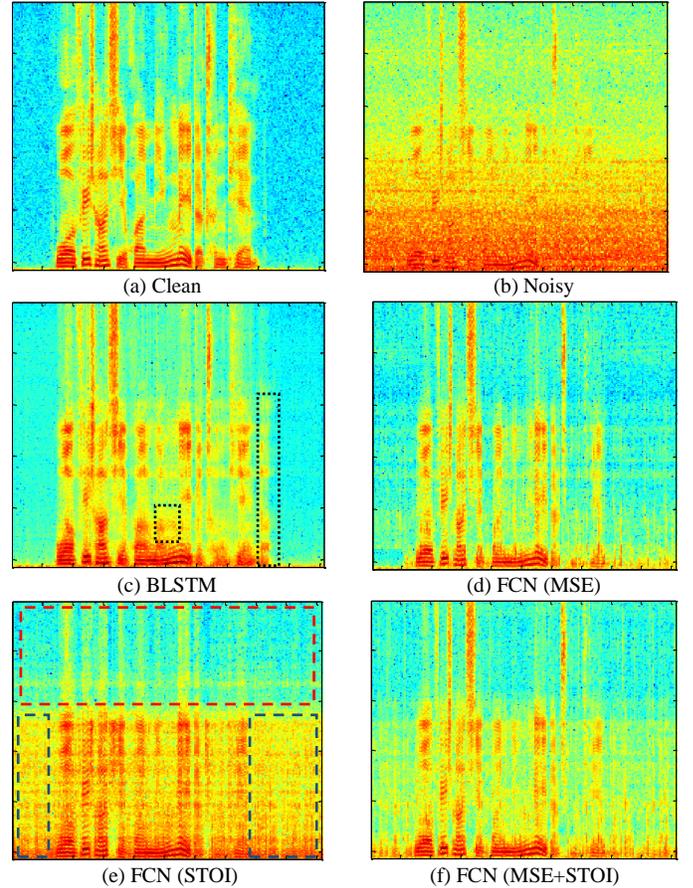

Fig. 9. Spectrograms of an MHINT utterance: (a) clean speech, (b) noisy speech (engine noise at -3 dB) (STOI= 0.6470, PESQ= 1.5558), (c) enhanced speech by BLSTM (STOI= 0.7677, PESQ= 1.7398), (d) enhanced speech by FCN with MSE objective function (STOI= 0.7764, PESQ = 1.8532), (e) enhanced speech by FCN with STOI objective function (STOI= **0.7958**, PESQ= 1.7191), and (f) enhanced speech by FCN with MSE+STOI objective function (STOI= 0.7860, PESQ = **1.8843**).

*3) Spectrogram Comparison*

Next, we present the spectrograms of a clean MHINT utterance, the same utterance corrupted by engine noise at -3 dB, and enhanced speeches by BLSTM and FCN with different objective functions in Fig. 9. Because the energy of speech components is less than that of noise, it is difficult to find out speech pattern in Fig. 9(b). Therefore, how to effectively recover the speech content for improving intelligibility is a critical concern in this case.

From Fig. 9(c), it can be observed that although BLSTM can most effectively remove the background noise, it misjudges the regions in the dashed black boxes as speech region. We found that this phenomenon usually happened when input noisy SNR is below 0dB and became much more severe in the -6dB case. This misjudgment may be due to the recurrent property in LSTM when noise energy is larger than speech. Next, when comparing Fig. 9(c) and (d), the speech components in FCN enhanced spectrogram seems to be more clear although there are some noise remains. This agrees with the results shown in Table II that FCN has higher STOI and lower PESQ scores compared to BLSTM. For STOI-optimized speech in Fig. 9(e), it can preserve much more (low- to mid-frequency) speech components when comparing to the noisy or MSE-optimized speech. However, because lacking definition about how to process high frequency parts (due to step 3 in the STOI evaluation and shown in the dashed brown box) and silent regions (due to step 1 in the STOI evaluation and shown in the dashed blue boxes), the optimized spectrogram looks noisy with high frequency components missing. Specifically, the missing high frequency components are attributed to the definition of STOI. As the highest one-third octave band (in step 3) has a center-frequency equal to ~4.3 kHz [34], the frequency components above this value do not affect the estimation of STOI (i.e., whether this region is very noisy or empty, the STOI value is not decreased). Therefore, FCN learns not to make any effort on this high-frequency region, and just removes most of the components. As pointed out previously, in addition to the silent regions being ignored, another reason caused noisy results comes from the calculation of intermediate intelligibility in (2), which is based on the correlation coefficient. Since the correlation coefficient is a scale- and shift-invariant measure, STOI just concerns the shape of (30-frames) temporal envelopes instead of their absolute positions. (i.e., when the vector is shifted or scaled by a constant, the correlation coefficient with another vector keeps unchanged). These two characteristics are the main reasons for decreased PESQ compared to the MSE-optimized counterpart. The two aforementioned phenomena of the STOI-optimized spectrogram can be mitigated by also incorporating MSE into the objective function, as shown in Fig. 9 (f).

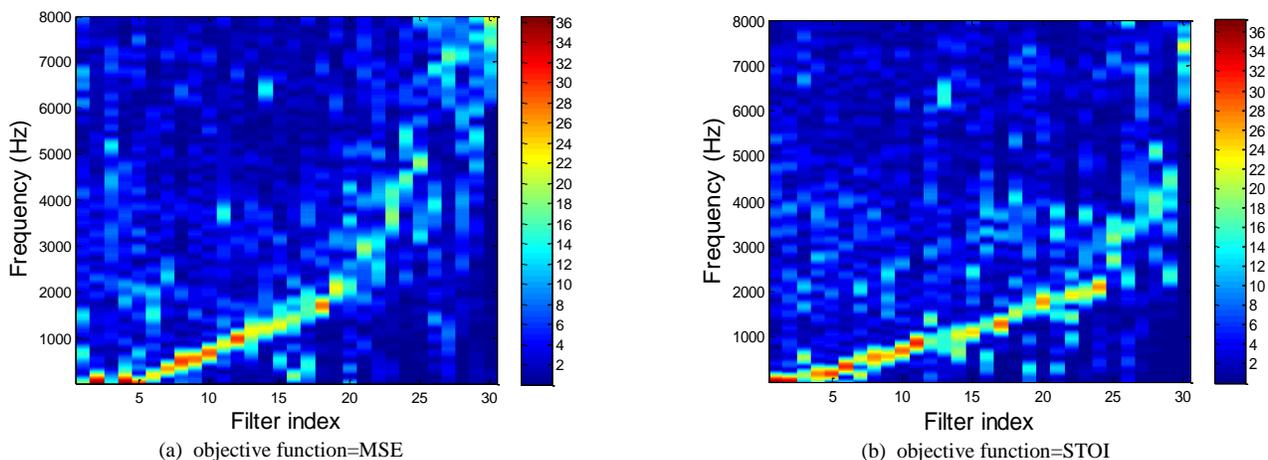

Fig. 10. Magnitude frequency response of the learned filters in the first layer of utterance-based FCN. The filter index is reordered by the location of the peak response for clear presentation. (a) Learned with the MSE objective function, and (b) learned with the STOI objective function.

*4) Analysis of Learned Filters*

In this section, we analyze the 30 learned filters in the first layer of FCN, and their magnitude frequency responses are illustrated in Fig. 10. Please note that the horizontal axis in the figure is the index of the filter, and we reordered the index according to the location of the peak response for clear presentation. From this figure, it can be observed that the pass-band of learned filters with MSE objective function (Fig. 10(a)) almost cover the entire frequency region (0–8 kHz). However, most of the pass-band of the STOI-optimized filters (Fig. 10(b)) concentrates on the frequency range below 4 kHz.

This may due to the high frequency components is not important for the estimation of STOI. In fact, the energy of the frequency region above 4 kHz occupies 31% of the entire range for the MSE-optimized filters. However, in the case of STOI-optimized filters, the ratio is only 21%, which implies that the high-frequency region is a stop-band for those filters. Therefore, this explains the missing high-frequency components in Fig. 9(e).

*5) Listening Test*

Although the intelligibility of noisy speech can be improved by denoising autoencoder for cochlear implant users [70, 71], this is usually not the case for speech evaluated on people with normal hearing [41, 42]. Therefore, the intelligibility improvement is still an open challenge even for deep learning-based enhancement methods [22]. This section sheds some light on the possible solutions and reports the listening test results of noisy, and FCN enhanced speech with different objective functions with real subjects. Twenty normal hearing native Mandarin Chinese subjects (sixteen males and four females) aged 23-45 participated in the listening tests. The same MHINT sentences used in the objective evaluations were adopted in the listening tests. Because real subjects were involved in this set of experiments, the number of test sets is confined to avoid biased results caused by listening fatigue [72] and ceiling effects of speech recognition [73]. Thus, we decided to prepare only two SNR levels (i.e., -3 and -6 dB), where intelligibility improvements are most needed in our test set. Each subject only participated in one SNR condition. In addition, we select the two more challenging noise types, namely engine and street noises, to form the test set.

The experiments were conducted in a quiet environment in which the background noise level was below 45 dB SPL. The stimuli were played to the subjects through a set of Sennheiser HD headphones at a comfortable listening level with our Speech-Evaluation-Toolkit (SET)[2]. Each subject participated in a total of 8 test conditions: 1 SNR levels × 2 noise types × 4 NR techniques—i.e., noisy, FCN (MSE), FCN (STOI), and FCN (MSE+STOI). Each condition contained ten sentences, and the order of the 8 conditions was randomized individually for each listener. None of the ten sentences was repeated across the test conditions. The subjects were instructed to verbally repeat what they heard and were allowed to repeat the stimuli twice. The word correct rate (WCR) is used as the evaluation metric for speech intelligibility, which is calculated by dividing the number of correctly identified words by the total number of words under each test condition. In addition to intelligibility, we also evaluated the speech quality by mean opinion score (MOS) tests. Specifically, after listening to each stimulus, the subjects were also asked to rate the quality of the stimulus in a five-point Likert scale score (1: Bad, 2: Poor, 3: Fair, 4: Good, 5: Excellent).

---

[2] Available at https://github.com/Dati1020/Speech-Evaluation-Toolkit

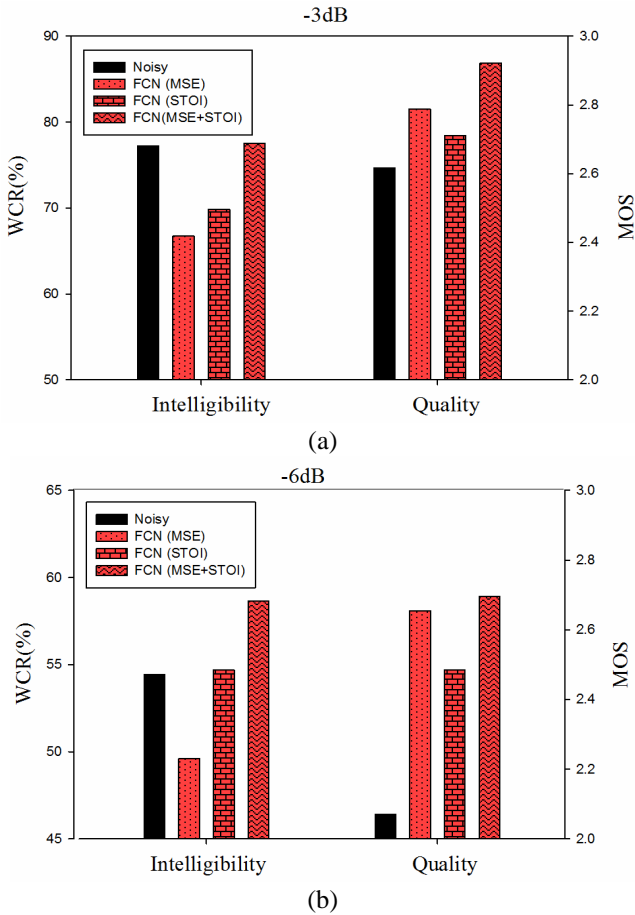

Fig. 11. Average WCR and MOS scores of human subjects for (a) -3dB and (b) -6dB.

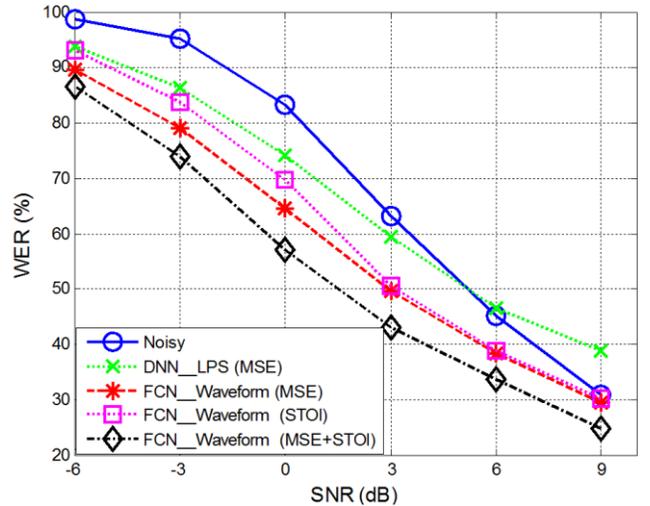

Fig. 12. WER of Google ASR for noisy speech, DNN-based LPS enhancement method, and (utterance-wise) FCN-based waveform enhancement models with different objective functions. (The WER for clean speech is 9.84%)

Figure 11 illustrates the results of listening test for -3 dB and -6dB. We can first observe that although the quality of all the enhanced speech can be improved compared to the noisy one, intelligibility is not easy to be improved. This verifies two things. 1) As stated in the Introduction Section, improving speech intelligibility is more challenging than enhancing quality [41, 42]. For example, the intelligibility of MSE-optimized speech is generally worse than noisy speech as reported in [22]. 2) Speech intelligibility and speech quality are different aspects of speech. They are related to each other, yet not necessarily equivalent [74]. Speech with poor quality can be highly intelligible [75] (e.g., only optimizing STOI), while on the other hand speech with high quality may be totally unintelligible [76] (e.g., only optimizing MSE). Although the quality of STOI-optimized speech is worse than MMSE-based one, its intelligibility is better. This implies that the intelligibility model defined in STOI is really helpful for perservering speech contents.

The results of optimizing MSE and STOI simultaneously seem to acquire advantages from the two terms, and hence can obtain the best performance in both intelligibility and quality. We also found that the intelligibility improvement in -3 dB SNR condition is very limited. This may be due to the fact that there is no much room for improvement since human ears are quite robust to moderate noises (WCR ~80% under this noisy condition). On the other hand, the intelligibility improvement is statistical significant ($p<0.05$) in the -6 dB SNR condition.

*6) ASR Experiments*

We have demonstrated that the proposed utterance-based FCN enhancement model can handle any kind of objective functions. To further confirm the applicability of the framework, we test the speech enhancement on the performance of ASR. Although the WER is widely used as an evaluation criterion, it is difficult to formulate the criterion in a specific objective function for enhancement optimization. Several studies have shown that speech enhancement can increase the noise-robustness of ASR [8, 9, 43, 77-82]. Some research [43, 44] has further shown that the CC between the improvement in WER of ASR and the improvement in STOI is higher than other objective evaluation scores (e.g., Moore et al. [36] showed that the CC can reach to 0.79). Since we demand high accuracy noise-robust ASR in real-world applications, a speech enhancement front-end which considers both MMSE and STOI may achieve better ASR performance than simply MMSE-optimized alternatives. Note that we are not pursuing a state-of-the-art noise-robust ASR system; instead we treat the ASR as an additional objective evaluation metric. In this study, we took a well-trained ASR (Google Speech Recognition) [83] to test speech recognition performance.

The same MHINT test sentences used in the objective evaluations were also adopted in the ASR experiment, and the results reported were averaged across the three noise types. The WER of ASR for noisy speech, enhanced speech by LPS-based DNN method, and waveform-based FCN enhancement models with different objective functions are shown in Fig. 12. This figure provides the following four observations: 1) the conventional DNN-based LPS enhancement method can only provide WER improvement under low-SNR conditions. Its WER is even worse than the noisy speech in the cases when SNR is higher than 6dB. 2) All the FCN enhanced speech samples can obtain lower WER compared to the noisy ones, and the improvement at around 0 dB is most obvious. 3) The WER of STOI-optimized speech is worse than that of MSE-optimized speech. This may be due to the spectrogram of STOI-optimized speech remaining too noisy for ASR (compare Fig. 9 (d) and (e)). Furthermore, PESQ is decreased by changing the objective function from MSE to STOI (compare the 8th to 11th columns in Table II). Although not as highly correlated as the STOI case, the decrease of PESQ may also degrade the ASR performance (the correlation coefficient between improvement in WER and the improvements in PESQ is 0.55 [43]). Therefore, most of the WER reduction from increasing STOI might be canceled out by the decreasing PESQ. 4) As the results of listening test, when incorporating both MSE and STOI into the objective function of FCN, the WER can be considerably reduced compared to the MSE-optimized model. This verifies that bringing STOI into objective function of speech enhancement can also help ASR to identify the speech content under noisy conditions.

Although this ASR experiment was tested on a trained system, this is indeed more practical in many real-world applications where an ASR engine is supplied by a third-party. Our proposed FCN enhancement model can simply be treated as pre-processing to obtain a more noise-robust ASR.

In summary, although optimizing STOI alone only provides marginal WER improvements, incorporating STOI with MSE as a new objective function can obtain considerable benefits. This again shows that the intelligibility model defined in STOI is helpful for persevering speech contents. However, because STOI does not consider non-speech regions and is based on CC in the original definition, its noise suppression ability is not enough for ASR applications. Therefore, optimizing STOI and MSE simultaneously seems to strike a good balance between noise reduction (by MSE term) and speech intelligibility improvement (by STOI term).

### C. Experiment on the CHiME-2 data set

Finally, we intend to test the proposed algorithm in a more challenging task. The noisy and reverberant CHiME2 dataset were adopted to evaluate the effect of removing both additive and convolutive noise simultaneously. The reverberant and noisy signals were created by first convolving the clean signals in the WSJ0-5k corpus with binaural room impulse responses (BRIRs), and then adding binaural recordings of genuine room noise at six different SNR levels linearly spaced from -6 dB to 9 dB SNR levels. The noises included a rich collection of sounds, such as children talking, electronic devices, distant noises, background music, and so on. There was a 7138-utterance training set (~14.5h in total), which included various noisy mixtures and speakers, a 2460 utterance development set (~4.5h in total), which was derived from 410 clean speech utterances, each mixed with a noise signal at six different noise levels, and an evaluation set, which included 1980 utterances (~4h in total) derived from 330 clean speech signals. The original clean utterances from the WSJ0-5k were used as the output targets.

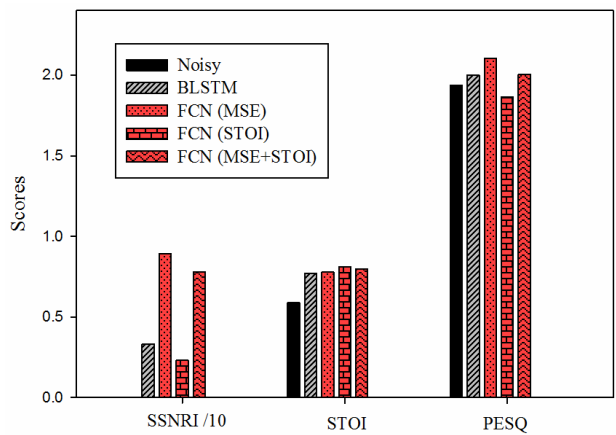

Fig. 13. Average objective evaluation scores for different models on the CHiME2 data set.

In this set of experiments, we used the same model structure as that used in the MHINT experiment. The optimal training epoch was decided by the development set. Fig. 13 illustrates the average objective evaluation scores for the different models. From these results, we can first observe that both the improvements of SSNR and PESQ are not so obvious compared to the MHINT experiment because of the appearance of convolutive noise. In addition, STOI optimization can also achieve the highest STOI score for reverberant speech. Overall, the performance trends of different models are similar to the previous MHINT experiment, except that the PESQ score of FCN (MSE) can also outperform BLSTM. Please note that the mathematical model (convolution) for producing reverberant speech is the same as single layer FCN without activation function. Therefore, FCN may be more suitable to model reverberation; nevertheless a more rigorous experiment is needed to verify this, which will be our future work.

### V. DISCUSSION

Our initial purpose in this study is to reduce the gap between the model optimization and evaluation criterions for deep learning based speech enhancement systems. Based on our proposed algorithm which takes the STOI as an optimization criterion, the system can indeed improve speech intelligibility. However, directly applying it as the only objective function seems to be not good enough. This is mainly because of that STOI does not define how silent and high frequency regions should be processed; therefore, the STOI optimized speech may appear in an unexpected way in these regions. Accordingly the objective function formed by combining MSE and STOI is a reasonable solution. As confirmed from the experimental results of listening tests and ASR, optimizing MSE and STOI simultaneously can obtain the best performance. In addition to the combination of these two terms, we also designed a conditional objective function, which assigns different loss in different regions. More specifically, to reduce the influence of the MSE term on the speech region, we only applied it in the silent regions instead of the whole utterance. Hence, the objective function can be represented as the following equation.

$$O = \begin{cases} \frac{\alpha}{|Si|}\|w_u(t) - \hat{w}_u(t)\|_2^2, & if\ t \in \text{silent of } w_u(t) \\ -stoi(w_u(t), \hat{w}_u(t)), & (if\ t \in \text{speech of } w_u(t)) \end{cases}, \quad (7)$$

where $|Si|$ is the number of sample points in silent regions. We put the second condition about STOI in parentheses, because this condition is already considered in the original STOI evaluation. Unfortunately, preliminary experimental results show that this conditional objective function does not work very well. Since the target of the MSE term, $w_u(t)$, is usually close to zero (silent region), the model only learns to scale down the weights (this would not degrade STOI term because it is based on CC). Therefore, the output utterance is a trivial solution similar to the STOI-optimized speech only with very small energy.

The calculation of STOI seems only depend on the magnitude spectrogram and is not related to phase (hence waveform-based model is not necessary). However, if we only focus on optimizing magnitude spectrogram, the magnitude spectrogram of the synthesized time-domain signal cannot keep optimality [47, 84]. Hence, the phase should also be considered in the optimization process or performing speech enhancement in the waveform-domain directly. In summary, although we adopted the STOI as the objective function, the model is optimized based on the difference of enhanced and clean target waveforms. Accordingly, the optimization process considers magnitude spectrum and phase simultaneously.

In [85], Kolbæk et al. applied DNNs to optimize approximate-STOI with several approximations on the original STOI definition. Possibly due to those approximations along with the limitation of a short segment-based model, their method could not outperform MSE-based systems. The present study, on the other hand, intends to directly optimize STOI without any approximation by using a FCN utterance-based model. The benefit of this utterance-based enhancement is that it can integrate the long-term speech continuity property (determined by the continuous vocal tract movement in producing continuous speech utterances). This continuity helps to improve the speech intelligibility which could not be explored in frame-based enhancement models even context features are also used as inputs [17].

As showed in Fig. 6, our proposed utterance-based waveform enhancement FCN model is flexible which can be easily extended to other diverse objective functions, from the local time scale (frame or short segment) to the global time scale (long segment or utterance), and from measures in the time domain to the frequency domain. The STOI optimization demonstrated in this paper is just one example. Specifically, the STOI function in Fig. 6 can be replaced by another specific evaluation metrics (e.g. SNR, SSNR or PESQ, etc.). When a new objective evaluation metric is proposed, our model can be readily applied to optimize the metrics, as long as every step in the evaluation metric is differentiable (otherwise, a continuous approximation function is needed).

Last but not least, the experimental results of listening tests and ASR confirm the importance of the objective function for optimizing the model parameters. Although the model structure is fixed, changing the objective functions may induce very different results. Currently, some evaluation metrics still not perfectly reflect the human perception while it is expectable that more accurate evaluation metrics will be proposed in the future. By combining the proposed framework with more accurate evaluation metrics, we hope the mismatch between the training objective and human perception can be effectively reduced.

## VI. CONCLUSION

This paper proposes a speech enhancement framework which takes testing evaluation metrics in model parameter training. This is different from conventional methods which takes un-consistent objectives in training and evaluations. In order to solve the mismatch problem, we proposed an end-to-end utterance-based raw waveform speech enhancement system by FCN architecture. Through the novel framework, several problems that exist in conventional DNN-based enhancement model can be solved simultaneously. 1) The mismatch between the true targets of speech enhancement and the employed objective function can be solved by utterance-based waveform optimization. 2) There is no need to map the time domain waveform to the frequency domain for enhancing the magnitude spectrogram. Therefore, all the related pre- and post-processing can be avoided. 3) Because the proposed model directly denoises the noisy waveform, the phase information is not ignored. 4) The discontinuity of enhanced speech observed in conventional frame-based processing is solved by treating each utterance as a whole. Since deep learning has a strong capacity to learn a mapping function, we found that it is extremely important to apply our real target as the objective function for optimization. The STOI optimization shows its excellent connections to the purpose of speech intelligibility improvements when it is formulated into objective functions. By efficiently integrating this type of objective functions in data-driven model learning, it is possible to reveal real connections of physical acoustic features with the complex perception quantities.

## REFERENCES

[1] Y. Xu, J. Du, L.-R. Dai, and C.-H. Lee, "An experimental study on speech enhancement based on deep neural networks," *IEEE Signal Processing Letters,* vol. 21, pp. 65-68, 2014.

[2] Y. Xu, J. Du, L.-R. Dai, and C.-H. Lee, "Global variance equalization for improving deep neural network based speech enhancement," in *IEEE China Summit & International Conference on Signal and Information Processing (ChinaSIP)*, 2014, pp. 71-75.

[3] Y. Xu, J. Du, L.-R. Dai, and C.-H. Lee, "Dynamic noise aware training for speech enhancement based on deep neural networks," in *INTERSPEECH*, 2014, pp. 2670-2674.

[4] Y. Xu, J. Du, L.-R. Dai, and C.-H. Lee, "A regression approach to

speech enhancement based on deep neural networks," *IEEE/ACM Transactions on Audio, Speech, and Language Processing* vol. 23, pp. 7-19, 2015.
[5] T. Gao, J. Du, L.-R. Dai, and C.-H. Lee, "SNR-based progressive learning of deep neural network for speech enhancement," in *INTERSPEECH*, 2016, pp. 3713-3717.
[6] S.-W. Fu, Y. Tsao, and X. Lu, "SNR-aware convolutional neural network modeling for speech enhancement," in *INTERSPEECH*, 2016, pp. 3768-3772.
[7] X. Lu, Y. Tsao, S. Matsuda, and C. Hori, "Speech enhancement based on deep denoising autoencoder," in *INTERSPEECH*, 2013, pp. 436-440.
[8] Z. Chen, S. Watanabe, H. Erdogan, and J. Hershey, "Integration of speech enhancement and recognition using long-short term memory recurrent neural network," in *INTERSPEECH*, 2015.
[9] F. Weninger, H. Erdogan, S. Watanabe, E. Vincent, J. Le Roux, J. R. Hershey*, et al.*, "Speech enhancement with LSTM recurrent neural networks and its application to noise-robust ASR," in *International Conference on Latent Variable Analysis and Signal Separation*, 2015, pp. 91-99.
[10] B. Xia and C. Bao, "Wiener filtering based speech enhancement with weighted denoising auto-encoder and noise classification," *Speech Communication,* vol. 60, pp. 13-29, 2014.
[11] P. G. Shivakumar and P. Georgiou, "Perception optimized deep denoising autoencoders for speech enhancement," in *INTERSPEECH*, 2016, pp. 3743-3747.
[12] D. S. Williamson, Y. Wang, and D. Wang, "Complex ratio masking for joint enhancement of magnitude and phase," in *Acoustics, Speech and Signal Processing (ICASSP), 2016 IEEE International Conference on*, 2016, pp. 5220-5224.
[13] D. S. Williamson, Y. Wang, and D. Wang, "Complex ratio masking for monaural speech separation," *IEEE/ACM Transactions on Audio, Speech, and Language Processing,* vol. 24, pp. 483-492, 2016.
[14] S.-W. Fu, T.-y. Hu, Y. Tsao, and X. Lu, "Complex spectrogram enhancement by convolutional neural network with multi-metrics learning," in *MLSP*, 2017.
[15] Y. Xu, J. Du, Z. Huang, L.-R. Dai, and C.-H. Lee, "Multi-objective learning and mask-based post-processing for deep neural network based speech enhancement," in *INTERSPEECH*, 2015.
[16] A. Ogawa, S. Seki, K. Kinoshita, M. Delcroix, T. Yoshioka, T. Nakatani*, et al.*, "Robust example search using bottleneck features for example-based speech enhancement," in *INTERSPEECH*, 2016, pp. 3733-3737.
[17] S.-S. Wang, H.-T. Hwang, Y.-H. Lai, Y. Tsao, X. Lu, H.-M. Wang*, et al.*, "Improving denoising auto-encoder based speech enhancement with the speech parameter generation algorithm," in *APSIPA*, 2015, pp. 365-369.
[18] D. Wang and J. Chen, "Supervised speech separation based on deep learning: an overview," *arXiv preprint arXiv:1708.07524,* 2017.
[19] S. Wang, K. Li, Z. Huang, S. M. Siniscalchi, and C.-H. Lee, "A transfer learning and progressive stacking approach to reducing deep model sizes with an application to speech enhancement," in *Acoustics, Speech and Signal Processing (ICASSP), 2017 IEEE International Conference on*, 2017, pp. 5575-5579.
[20] D. Michelsanti and Z.-H. Tan, "Conditional generative adversarial networks for speech enhancement and noise-robust speaker verification," in *INTERSPEECH*, 2017, pp. 2008-2012.
[21] P.-S. Huang, M. Kim, M. Hasegawa-Johnson, and P. Smaragdis, "Joint optimization of masks and deep recurrent neural networks for monaural source separation," *IEEE/ACM Transactions on Audio, Speech and Language Processing (TASLP),* vol. 23, pp. 2136-2147, 2015.
[22] M. Kolbæk, Z.-H. Tan, and J. Jensen, "Speech intelligibility potential of general and specialized deep neural network based speech enhancement systems," *IEEE/ACM Transactions on Audio, Speech, and Language Processing,* vol. 25, pp. 153-167, 2017.
[23] E. M. Grais and M. D. Plumbley, "Single channel audio source separation using convolutional denoising autoencoders," *arXiv preprint arXiv:1703.08019,* 2017.
[24] F. Weninger, J. R. Hershey, J. Le Roux, and B. Schuller, "Discriminatively trained recurrent neural networks for single-channel speech separation," in *IEEE Global Conference on Signal and Information Processing (GlobalSIP),*, 2014, pp. 577-581.
[25] Z. Chen, S. Watanabe, H. Erdogan, and J. R. Hershey, "Speech enhancement and recognition using multi-task learning of long short-term memory recurrent neural networks," in *INTERSPEECH*, 2015.
[26] H. Erdogan, J. R. Hershey, S. Watanabe, and J. Le Roux, "Phase-sensitive and recognition-boosted speech separation using deep recurrent neural networks," in *2015 IEEE International Conference on Acoustics, Speech and Signal Processing (ICASSP)*, 2015, pp. 708-712.
[27] H. Erdogan, J. R. Hershey, S. Watanabe, and J. Le Roux, "Deep recurrent networks for separation and recognition of single-channel speech in nonstationary background audio," in *New Era for Robust Speech Recognition*, ed: Springer, 2017, pp. 165-186.
[28] L. Sun, J. Du, L.-R. Dai, and C.-H. Lee, "Multiple-target deep learning for LSTM-RNN based speech enhancement," in *Hands-free Speech Communications and Microphone Arrays (HSCMA)*, 2017, pp. 136-140.
[29] E. M. Grais, G. Roma, A. J. Simpson, and M. D. Plumbley, "Two-stage single-channel audio source separation using deep neural networks," *IEEE/ACM Transactions on Audio, Speech, and Language Processing,* vol. 25, pp. 1773-1783, 2017.
[30] E. M. Grais, H. Wierstorf, D. Ward, and M. D. Plumbley, "Multi-resolution fully convolutional neural networks for monaural audio source separation," *arXiv preprint arXiv:1710.11473,* 2017.
[31] Y. Ephraim and D. Malah, "Speech enhancement using a minimum-mean square error short-time spectral amplitude estimator," *IEEE Transactions on Acoustics, Speech and Signal Processing,* vol. 32, pp. 1109-1121, 1984.
[32] J. Benesty, S. Makino, and J. D. Chen, *Speech enhancement* Springer, 2005.
[33] A. Rix, J. Beerends, M. Hollier, and A. Hekstra, "Perceptual evaluation of speech quality (PESQ), an objective method for end-to-end speech quality assessment of narrowband telephone networks and speech codecs," *ITU-T Recommendation,* p. 862, 2001.
[34] C. H. Taal, R. C. Hendriks, R. Heusdens, and J. Jensen, "An algorithm for intelligibility prediction of time–frequency weighted noisy speech," *IEEE Transactions on Audio, Speech, and Language Processing,* vol. 19, pp. 2125-2136, 2011.
[35] J. R. Hershey, Z. Chen, J. Le Roux, and S. Watanabe, "Deep clustering: Discriminative embeddings for segmentation and separation," in *Acoustics, Speech and Signal Processing (ICASSP), 2016 IEEE International Conference on*, 2016, pp. 31-35.
[36] J. Long, E. Shelhamer, and T. Darrell, "Fully convolutional networks for semantic segmentation," in *Proceedings of the IEEE Conference on Computer Vision and Pattern Recognition*, 2015, pp. 3431-3440.
[37] S.-W. Fu, Y. Tsao, X. Lu, and H. Kawai, "Raw waveform-based speech enhancement by fully convolutional networks," in *APSIPA*, 2017.
[38] M. Kolbæk, D. Yu, Z.-H. Tan, and J. Jensen, "Multi-talker speech separation with utterance-level permutation invariant training of deep recurrent neural networks," *IEEE/ACM Transactions on Audio, Speech, and Language Processing,* 2017.
[39] D. P. Bertsekas, "Nondifferentiable optimization via approximation," *Nondifferentiable optimization,* pp. 1-25, 1975.
[40] Y. Koizumi, K. Niwa, Y. Hioka, K. Kobayashi, and Y. Haneda, "DNN-based source enhancement self-optimized by reinforcement learning using sound quality measurements," in *Acoustics, Speech and Signal Processing (ICASSP), 2017 IEEE International Conference on*, 2017, pp. 81-85.
[41] P. C. Loizou and G. Kim, "Reasons why current speech-enhancement algorithms do not improve speech intelligibility and suggested solutions," *IEEE transactions on audio, speech, and language processing,* vol. 19, pp. 47-56, 2011.
[42] P. C. Loizou, *Speech enhancement: theory and practice*: CRC press, 2013.
[43] A. Moore, P. P. Parada, and P. Naylor, "Speech enhancement for robust automatic speech recognition: Evaluation using a baseline system and instrumental measures," *Computer Speech & Language,* 2016.
[44] D. A. Thomsen and C. E. Andersen, "Speech enhancement and noise-robust automatic speech recognition," Aalborg University, 2015.
[45] B. Lecouteux, M. Vacher, and F. Portet, "Distant speech recognition

for home automation: Preliminary experimental results in a smart home," in *2011 6th Conference on Speech Technology and Human-Computer Dialogue (SpeD)*, 2011, pp. 1-10.

[46] K. Paliwal, K. Wójcicki, and B. Shannon, "The importance of phase in speech enhancement," *speech communication*, vol. 53, pp. 465-494, 2011.

[47] J. Le Roux, "Phase-controlled sound transfer based on maximally-inconsistent spectrograms," *Signal*, vol. 5, p. 10, 2011.

[48] S. Pascual, A. Bonafonte, and J. Serrà, "SEGAN: Speech enhancement generative adversarial network," *arXiv preprint arXiv:1703.09452*, 2017.

[49] K. Qian, Y. Zhang, S. Chang, X. Yang, D. Florêncio, and M. Hasegawa-Johnson, "Speech enhancement using Bayesian Wavenet," in *INTERSPEECH*, 2017, pp. 2013-2017.

[50] D. Rethage, J. Pons, and X. Serra, "A Wavenet for speech denoising," *arXiv preprint arXiv:1706.07162*, 2017.

[51] A. V. Oppenheim, *Discrete-time signal processing*: Pearson Education India, 1999.

[52] A. v. d. Oord, S. Dieleman, H. Zen, K. Simonyan, O. Vinyals, A. Graves*, et al.*, "Wavenet: a generative model for raw audio," *arXiv preprint arXiv:1609.03499*, 2016.

[53] J. Bang-Jensen, G. Gutin, and A. Yeo, "When the greedy algorithm fails," *Discrete Optimization*, vol. 1, pp. 121-127, 2004.

[54] C. H. Taal, R. C. Hendriks, and R. Heusdens, "Speech energy redistribution for intelligibility improvement in noise based on a perceptual distortion measure," *Computer Speech & Language*, vol. 28, pp. 858-872, 2014.

[55] W. B. Kleijn, J. B. Crespo, R. C. Hendriks, P. Petkov, B. Sauert, and P. Vary, "Optimizing speech intelligibility in a noisy environment: A unified view," *IEEE Signal Processing Magazine*, vol. 32, pp. 43-54, 2015.

[56] S. Khademi, R. C. Hendriks, and W. B. Kleijn, "Intelligibility enhancement based on mutual information," *IEEE/ACM Transactions on Audio, Speech, and Language Processing*, vol. 25, pp. 1694-1708, 2017.

[57] L. Prechelt, "Early stopping-but when?," in *Neural Networks: Tricks of the trade*, ed: Springer, 1998, pp. 55-69.

[58] F. Chollet. (2015). *Keras*. Available: https://github.com/fchollet/keras

[59] T. T. D. Team, R. Al-Rfou, G. Alain, A. Almahairi, C. Angermueller, D. Bahdanau*, et al.*, "Theano: A Python framework for fast computation of mathematical expressions," *arXiv preprint arXiv:1605.02688*, 2016.

[60] J. W. Lyons, "DARPA TIMIT acoustic-phonetic continuous speech corpus," *National Institute of Standards and Technology*, 1993.

[61] L. L. Wong, S. D. Soli, S. Liu, N. Han, and M.-W. Huang, "Development of the Mandarin hearing in noise test (MHINT)," *Ear and hearing*, vol. 28, pp. 70S-74S, 2007.

[62] E. Vincent, J. Barker, S. Watanabe, J. Le Roux, F. Nesta, and M. Matassoni, "The second 'CHiME' speech separation and recognition challenge: Datasets, tasks and baselines," in *Acoustics, Speech and Signal Processing (ICASSP), 2013 IEEE International Conference on*, 2013, pp. 126-130.

[63] G. Hu. *100 nonspeech environmental sounds, 2004 [Online]*. Available: http://www.cse.ohio-state.edu/pnl/corpus/HuCorpus.html.

[64] A. L. Maas, A. Y. Hannun, and A. Y. Ng, "Rectifier nonlinearities improve neural network acoustic models," in *Proc. ICML*, 2013.

[65] D. Kingma and J. Ba, "Adam: A method for stochastic optimization," *arXiv preprint arXiv:1412.6980*, 2014.

[66] S. Ioffe and C. Szegedy, "Batch normalization: Accelerating deep network training by reducing internal covariate shift," in *Proceedings of the 32nd International Conference on Machine Learning (ICML-15)*, 2015, pp. 448-456.

[67] G. Hinton, N. Srivastava, and K. Swersky, "RMSProp: Divide the gradient by a running average of its recent magnitude," *Neural networks for machine learning, Coursera lecture 6e*, 2012.

[68] J. Chen, J. Benesty, Y. Huang, and E. Diethorn, "Fundamentals of noise reduction in spring handbook of speech processing," ed: Springer, 2008.

[69] Y. Li and D. Wang, "On the optimality of ideal binary time–frequency masks," *Speech Communication*, vol. 51, pp. 230-239, 2009.

[70] Y.-H. Lai, F. Chen, S.-S. Wang, X. Lu, Y. Tsao, and C.-H. Lee, "A deep denoising autoencoder approach to improving the intelligibility of vocoded speech in cochlear implant simulation," *IEEE Transactions on Biomedical Engineering*, vol. 64, pp. 1568-1578, 2017.

[71] S.-S. Wang, Y. Tsao, H.-L. S. Wang, Y.-H. Lai, and L. P.-H. Li, "A deep learning based noise reduction approach to improve speech intelligibility for cochlear implant recipients in the presence of competing speech noise," in *APSIPA*, 2017.

[72] C. B. Hicks and A. M. Tharpe, "Listening effort and fatigue in school-age children with and without hearing loss," *Journal of Speech, Language, and Hearing Research*, vol. 45, pp. 573-584, 2002.

[73] R. H. Gifford, J. K. Shallop, and A. M. Peterson, "Speech recognition materials and ceiling effects: Considerations for cochlear implant programs," *Audiology and Neurotology*, vol. 13, pp. 193-205, 2008.

[74] D. Li, "Deep neural network approach for single channel speech enhancement processing," Université d'Ottawa/University of Ottawa, 2016.

[75] R. V. Shannon, F.-G. Zeng, V. Kamath, J. Wygonski, and M. Ekelid, "Speech recognition with primarily temporal cues," *Science*, vol. 270, p. 303, 1995.

[76] J. H. James, B. Chen, and L. Garrison, "Implementing VoIP: a voice transmission performance progress report," *IEEE Communications Magazine*, vol. 42, pp. 36-41, 2004.

[77] D. Baby, J. F. Gemmeke, and T. Virtanen, "Exemplar-based speech enhancement for deep neural network based automatic speech recognition," in *Acoustics, Speech and Signal Processing (ICASSP), 2015 IEEE International Conference on*, 2015, pp. 4485-4489.

[78] R. F. Astudillo, J. Correia, and I. Trancoso, "Integration of DNN based speech enhancement and ASR," in *INTERSPEECH*, 2015.

[79] N. Yoma, F. McInnes, and M. Jack, "Lateral inhibition net and weighted matching algorithms for speech recognition in noise," *IEE Proceedings-Vision, Image and Signal Processing*, vol. 143, pp. 324-330, 1996.

[80] C. Donahue, B. Li, and R. Prabhavalkar, "Exploring speech enhancement with generative adversarial networks for robust speech recognition," *arXiv preprint arXiv:1711.05747*, 2017.

[81] T. Ochiai, S. Watanabe, T. Hori, and J. R. Hershey, "Multichannel end-to-end speech recognition," in *International Conference on Machine Learing (ICML)*, 2017.

[82] T. Ochiai, S. Watanabe, and S. Katagiri, "Does speech enhancement work with end-to-end ASR objectives?: Experimental analysis of multichannel end-to-end ASR," in *MLSP*, 2017.

[83] A. Zhang. (2017). *Speech Recognition (Version 3.6) [Software]*. Available: https://github.com/Uberi/speech_recognition#readme.

[84] T. Gerkmann, M. Krawczyk-Becker, and J. Le Roux, "Phase processing for single-channel speech enhancement: history and recent advances," *IEEE Signal Processing Magazine*, vol. 32, pp. 55-66, 2015.

[85] M. Kolbæk, Z.-H. Tan, and J. Jensen, "Monaural speech enhancement using deep neural networks by maximizing a short-time objective intelligibility measure," *arXiv preprint arXiv:1802.00604*, 2018.